\documentclass[conference]{IEEEtran}
\IEEEoverridecommandlockouts
\usepackage{cite}
\usepackage{amsmath,amssymb,amsfonts}
\usepackage{algorithmic}
\usepackage{graphicx}
\usepackage{textcomp}
\usepackage{xcolor}
\usepackage{capt-of}  
\usepackage{cuted} 
\def\BibTeX{{\rm B\kern-.05em{\sc i\kern-.025em b}\kern-.08em
    T\kern-.1667em\lower.7ex\hbox{E}\kern-.125emX}}
\begin{document}

\title{Wildfire danger prediction optimization with transfer learning\\
}

\author{
\IEEEauthorblockN{1\textsuperscript{st} Spiros Maggioros}
\IEEEauthorblockA{\textit{School of Electrical \& Computer Engineering} \\
\textit{National Technical University of Athens}\\
Athens, Greece \\
spirosmag@ieee.org}
\and
\IEEEauthorblockN{2\textsuperscript{nd} Nikos Tsalkitzis}
\IEEEauthorblockA{\textit{School of Electrical \& Computer Engineering} \\
\textit{National Technical University of Athens}\\
Athens, Greece \\
nikostsalkitzhs@gmail.com}
}

\maketitle

\begin{abstract}
Convolutional Neural Networks (CNNs) have proven instrumental across various computer science domains, enabling advancements in object detection, classification, and anomaly detection. This paper explores the application of CNNs to analyze geospatial data specifically for identifying wildfire-affected areas. Leveraging transfer learning techniques, we fine-tuned CNN hyperparameters and integrated the Canadian Fire Weather Index (FWI) to assess moisture conditions. The study establishes a methodology for computing wildfire risk levels on a scale of 0 to 5, dynamically linked to weather patterns. Notably, through the integration of transfer learning, the CNN model achieved an impressive accuracy of 95\% in identifying burnt areas. This research sheds light on the inner workings of CNNs and their practical, real-time utility in predicting and mitigating wildfires. By combining transfer learning and CNNs, this study contributes a robust approach to assess burnt areas, facilitating timely interventions and preventative measures against conflagrations.
\end{abstract}


\section{Introduction}
In 2023, Europe saw the largest fire episode since 2000. Over 118,084 ha have burnt in 2023 (close to 27\% of the burnt area) is within Natura 2000 protected areas. Countries like Greece and Italy faced destructive wildfires over the summer, resulting in deaths, evacuations, destructive floods in winter and thousands of damaged properties.
Forest ecosystems are also related to the biochemical and global carbon cycle, thus, damaging them leads to unstable biological, physical and chemical properties of this ecosystem \cite{b1 , b2 , b4}.\\
Most studies of danger and fire prediction are aiming in meteorological data analysis, suffering from very low-accurate and non-reliable models.This is also a sign that the majority of those wildfires are caused by arsons, since models trained on Greek Wildfires had shown that meteorological data and wildfire incidents have extremely low correlation\cite{b1}.Thus, prediction of wildfires is a challenge for newer days and only live image analysis can be a reliable and high accurate model to help the already existing FWI prediction system(Fig. 1)
\cite{b5}.\\
\begin{figure}[htpb]
\centerline{\includegraphics[width=0.45\textwidth]{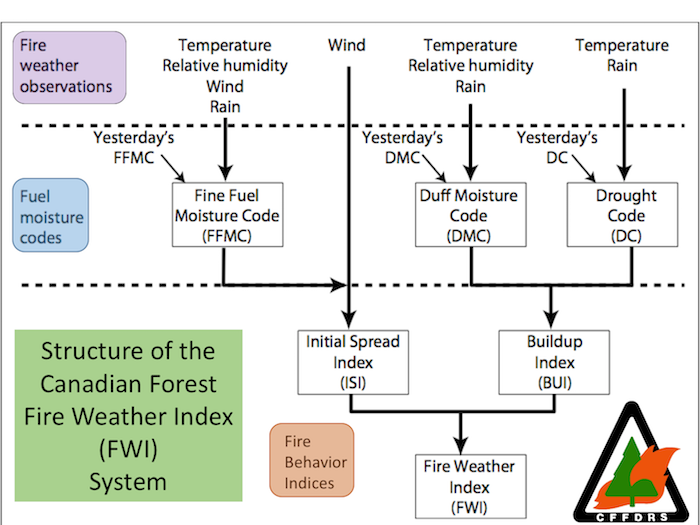}}
\caption{Canadian FWI system.}
\label{fig}
\end{figure}
Most models compute the outcome based on Temperature, Relative Humidity, Wind and Rain.A huge drawback to that design is not taking into account the short-history of the forest, giving previously destroyed forests a very high wildfire danger.Our model can precisely detect the short-history of a forest and in combination with the current FWI prediction system, is capable of producing very accurate results.

\section{Related Work}
In collaboration with our research team, numerous researchers have harnessed Convolutional Neural Networks (CNNs) not solely for general-purpose image classification, such as information extraction from social media \cite{b8}, but also for environmental examination. Geospatial data, for instance, has been instrumental in predicting natural disasters like earthquakes and wildfires \cite{b7}, as well as human-induced events like deforestation \cite{b10}. Given the diverse applications of CNNs in these domains, we advocate for comprehensive surveys related to the natural world.As our planet confronts an array of challenges, from excessive winter rainstorms to summer wildfires.\\
Spanning from satellite imagery \cite{b9} to geographic data, contemporary projects hold the potential to aid in preventing and protecting against these phenomena, while also uncovering intricate patterns that influence the scale of their consequences. From a computational perspective, an increase in surveys focusing on environmental imagery will contribute to higher model accuracy, leading to the development of more efficient algorithms tailored for this specific data, such as TorchGeo.

\section{Analysis on Greek Wildfires}
First and foremost, we will analyze the model we created, trained on Greek wildfires in 2018.We gathered all the incidents of wildfires and imputed the exact temperature, relative humidity, rain and wind values to the dataset using a Weather API.Note that we also have the destroyed area from the wildfire.In Fig. 2, we may inspect that the correlation of meteorological data with the destroyed area reveals no discernible linearity and indicates insufficient strength for the development of a reliable model.This led models to be extremely inaccurate in predicting the final danger level$(0-5)$.\\
We tried multiple regression algorithms as well as some optimized neural networks to predict the FWI values, yielding an outcome that does not align with the intended expectation.The algorithm that gave us the best results was the Random Forest(RF) regressor, an
algorithm that is well known for being used in meteorological data\cite{b6}. For the error function we used the Mean Absolute Error(MAE) :
\begin{equation}
MAE = \frac{1}{N} \cdot \sum_{i=1}^{N} |y_i - \hat{y_i}|
\end{equation}
where $N$ is the number of instances, $y_i$ is the real value and $\hat{y_i}$ is the predicted 
value.

\begin{table}[htbp]
\caption{Mean Absolute Errors of Regression algorithms}
\begin{center}
\begin{tabular}{|c|c|c|c|c|}
\hline
\textbf{Trained}&\multicolumn{4}{|c|}{\textbf{Mean Absolute Error}} \\
\cline{2-5} 
\textbf{Regressor} & \textbf{\textit{FFMC}}& \textbf{\textit{DMC}}&
\textbf{\textit{DC}}& \textbf{\textit{ISI}} \\
\hline
RF& $1.53$&$31.84$ &$104.5$  &$3.153$ \\
\hline
KNN& $1.70$ & $38.723$ & $133.90$ &$3.259$ \\
\hline
SVM& $1.5248$ & $42.92$  & $172.3$&$3.10$ \\
\hline
NuSVR& $1.50$& $43.0$ & $192.0 $& $3.1463$\\
\hline
\end{tabular}
\label{tab1}
\end{center}
\end{table}

For every implementation, the data was scaled for better results.We identify that the MAE is too high, thus, those models will perform poorly.The extremely high temperatures in Greece - in contrast with the more standard temperatures of the dataset - over the summer season, in combination with the small FWI dataset, led to these adverse outcomes.

\begin{figure}[htpb]
\centerline{\includegraphics[width=0.45\textwidth]{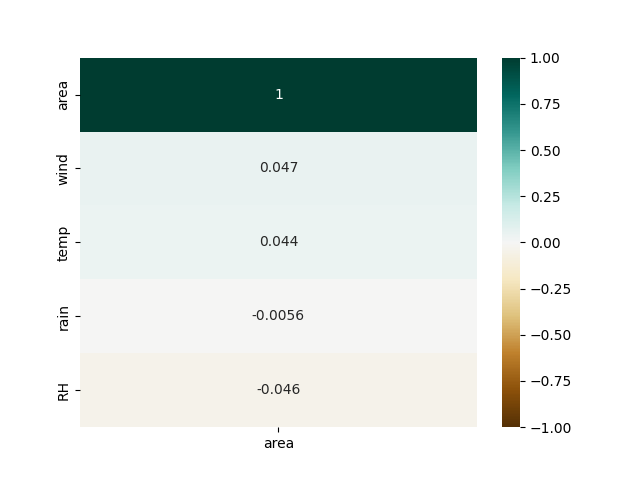}}
\caption{Correlation of meteorological data with destroyed area.}
\label{fig2}
\end{figure}

\section{Dataset}
We used a dataset from the Government and Municipalities of Québec \cite{b11} that is mainly used for research needs, such as analyzing the impact of climate change and post-fire regeneration.The dataset contains satellite images and aerial photographs from wildfires(Fig. 3) that occured in southern Québec(Fig. 4) from $1976$-present.
\begin{figure}[htpb]
\centerline{\includegraphics[width=0.45\textwidth]{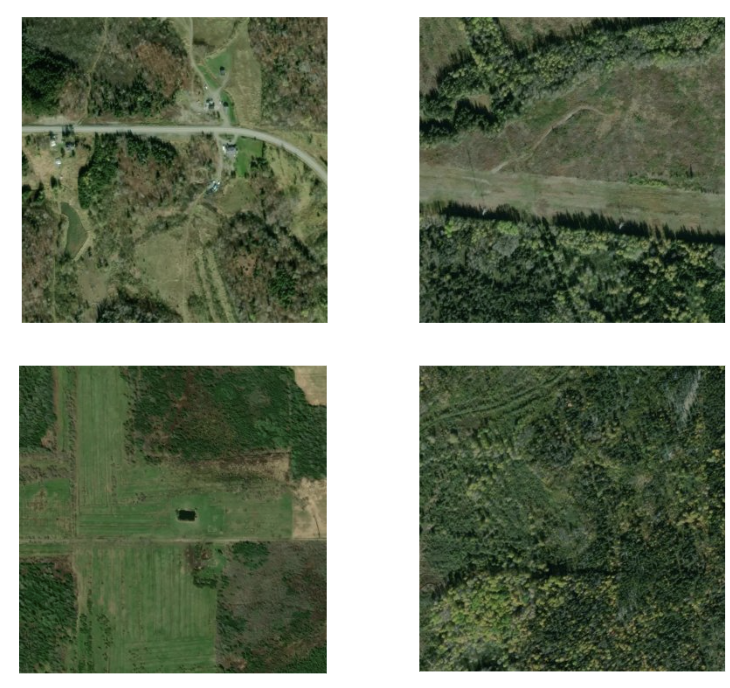}}
\caption{Satellite and aerial images from our dataset.}
\label{fig3}
\end{figure}
Satellite images of the exact areas have been extracted and imputed in the dataset using the MapBox API so we can have a more convenient format for deep learning.The final dataset contains $42850$ $350x350$ .png images into two classes
\begin{equation}
    Classes=
    \begin{cases}
      \text{Wildfire:} &\text{22710 images} \\
      \text{NoWildfire:} &\text{20140 images}
    \end{cases}
  \end{equation}

The dataset is finally splitted into 3 parts, Training($70\%$), Testing($15\%$) and Validation($15\%$).

\begin{figure}[htpb]
\centerline{\includegraphics[width=0.45\textwidth]{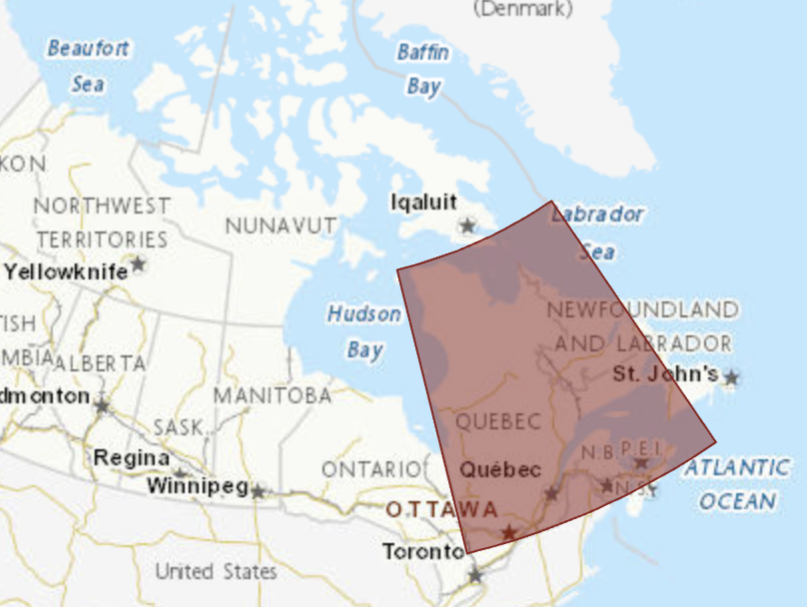}}
\caption{Mapping of forest fire incidents in Québec Canada.}
\label{fig4}
\end{figure}

\section{Theory, Evaluation and Optimization}

\subsection{Theory}
In modern society, deep learning is an emerging field with various applications.A sub-category is Convolutional Neural Networks which are radically implemented in image recognition, object detection, natural language processing and many more fields \cite{b13,b14,b15,b16} . As shown by the name of the algorithm, the whole process is based on the mathematical concept of convolution which is depicted in its discrete form: 
\begin{equation}
y[n] = x[n] * h[n] = \sum_{k=-\infty}^{\infty} x[k] \cdot h[n - k]
\end{equation}
We often use convolution in more than one axis, for example if the input $I$ is a 2D image then we have to use a 2D kernel K:
\begin{equation}
S(i,j) = (I * K)(i,j) = \sum_{m} \sum_{n} I(m,n) \cdot K(i-m,j-n)
\end{equation}
In image recognition, we examine every pixel of an image in RGB scale and we flatten it as an vectorized input in our model. If we have 28x28 images for just one layer we would have to calculate(with ANNs) 784x784=614.656 parameters(the size is increased proportionally to the number of hidden layers). So the first solution which the CNNs offer is that by looking just a segment of the image and not globally, some synapses drop out and the
parameters are decreasing \cite{b13,b14}. Practically, the network investigates regional patterns which makes it less vulnerable to spins of the images \cite{b15}.

\begin{figure}[htpb]
\centerline{\includegraphics[width=0.50\textwidth]{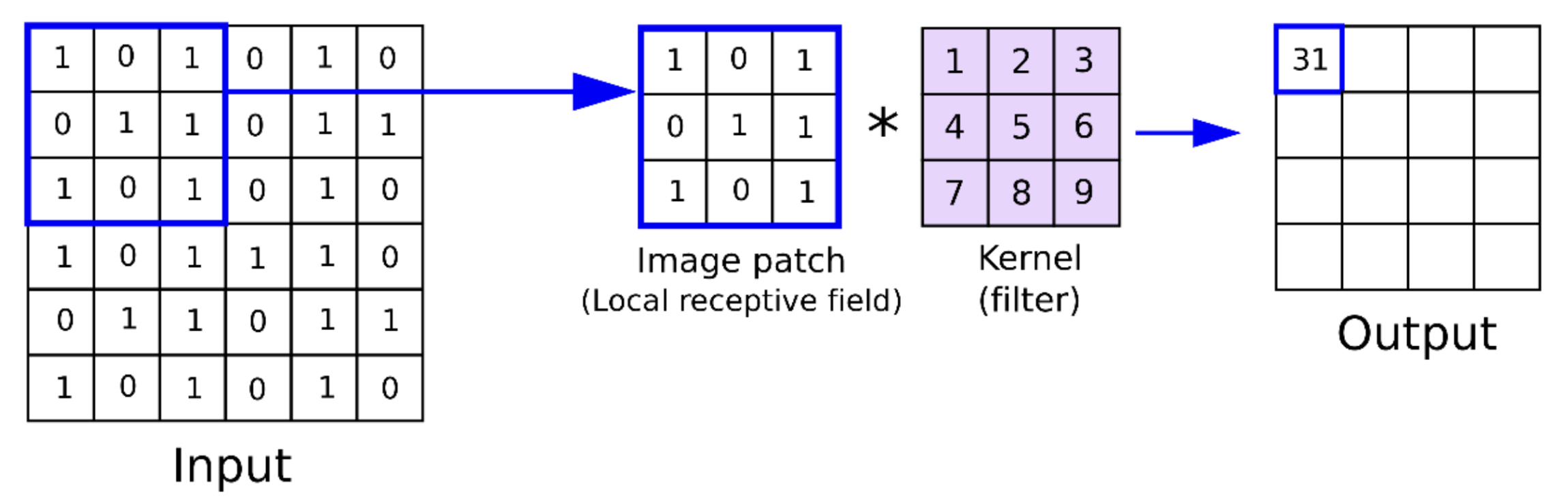}}
\caption{Image filtering using CNNs.}
\label{fig5}
\end{figure}
In Fig. 5 we see how the convolution layer works \cite{b12,b13}. In order to find some patterns(i.e edges or curves if we examine digits) we slide through the areas of the image the kernel-filter which declares the configuration we search for\cite{b12,b16}.\\
The output of this layer is called feature map and depicts if a specific pattern is present at the area we analyze. With this in mind, at every repetition of the training the model tries to learn the patterns of each class of images\cite{b15}. The calculation of the filter size required to attain the desired output can be achieved using the following formula: 
\begin{equation}
n_{L + 1} = \frac{n_L + 2 \cdot p - f}{s} + 1 
\end{equation}
Where $n_{L + 1}$ represents the height and width of the output, $n_{L}$ represents the height and width of the input, $p$ is the padding, $s$ the stride and finally, $f$ is the height and width of the window we want to use for the max-pooling.Solving for $f$ we get:
\begin{equation}
f = n_L + 2 \cdot p - s(n_{L + 1} - 1)
\end{equation}
In the above equation we can see two new ideas: padding and stride. The last one refers to how quickly the filter is applied to the input image(how many columns it covers until its next move)\cite{b12,b16}.By increasing the stride we eventually end up in diminished feature map \cite{b14}.Moreover the padding is used to grab the patterns that are located at the boarder of the images, by adding some extra pixels at the margins of the input. 
\begin{figure}[htpb]
\centerline{\includegraphics[width=0.50\textwidth]{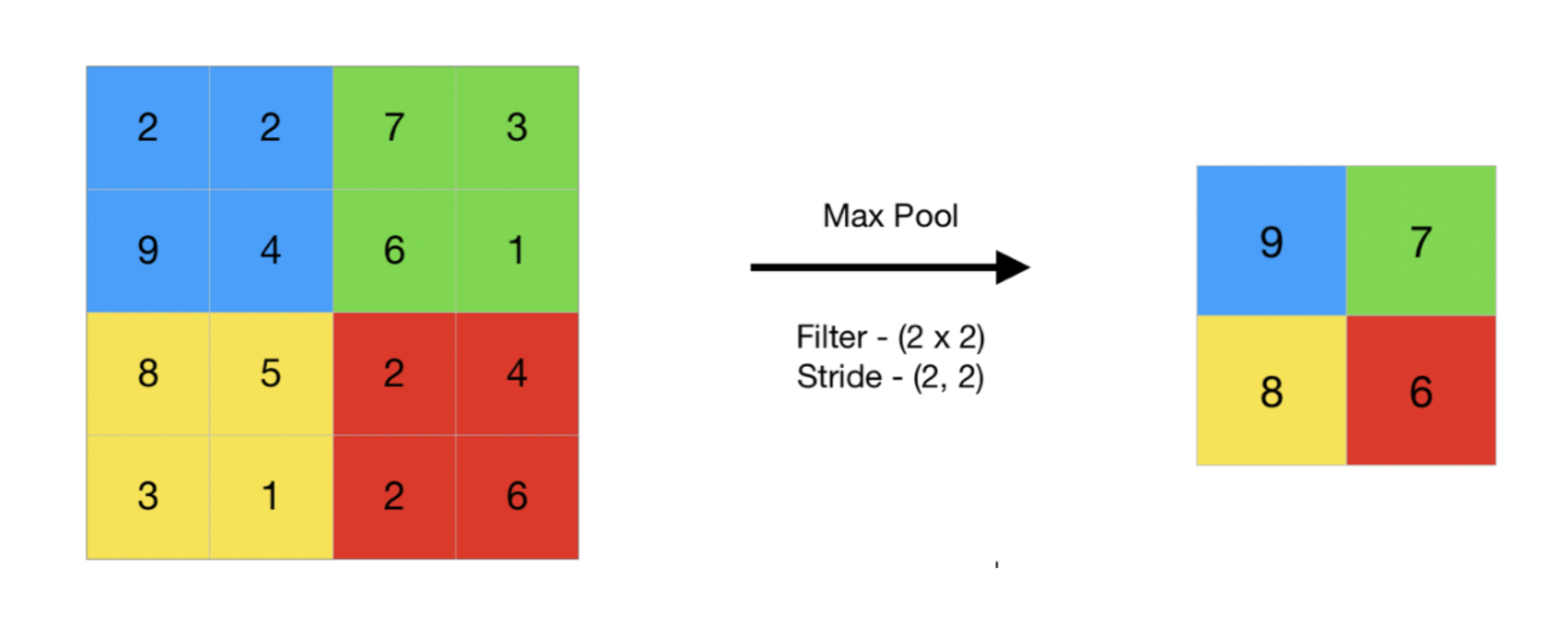}}
\caption{Pooling layer.}
\label{fig6}
\end{figure}
The next layer is named Pooling Layer(Fig. 6) \cite{b12, b14, b16} and it's responsible for reducing the spatial dimension of data which leads to faster implementation \cite{b15}.The feature map is significantly smaller than the initial one and with max pooling we keep only the most important features of it \cite{b13}.There also some other kinds of pooling(stochastic, gradient,min), the selection or which depends on the data we are called to analyze\cite{b14}.At the last stage of the above architecture, we generally use Softmax or Sigmoid as an activation function since we want to decide in which of K classes(or 2 classes like our example) the sample belongs to.There exists other options as well, like Leaky ReLU or Tanh that are not ideal for our problem(Fig. 7).

\begin{figure}[htpb]
\centerline{\includegraphics[width=0.50\textwidth]{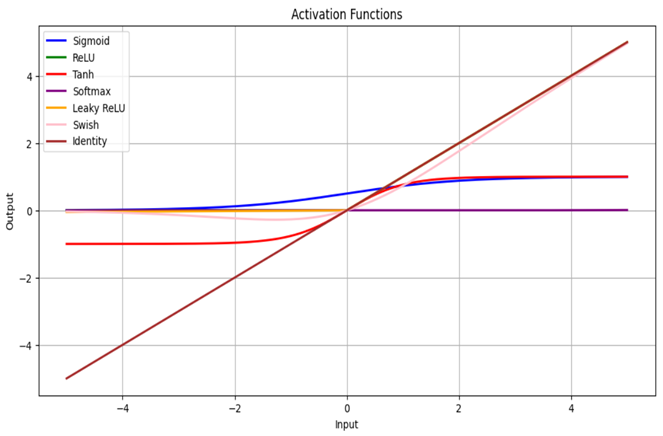}}
\caption{Options for activation functions.}
\label{fig7}
\end{figure}

\subsection{Experimental Results}
We will now explain the steps and the models we applied for the classification.For this task, we took advantage of Transfer Learning using pre-trained models of keras.All the models we tested were pre-trained on the ImageNet dataset, which contains multiple images including geospatial photographs.With this technique, we saved computational time and also managed to achieve very high accuracy on detecting burned areas \cite{b17}.Some of the models we implemented are: VGG16, VGG19, ResNet50, ResNetV2, Xception, EfficientNetB7, EfficientNetV2L, and NasNetLarge.\\
Before running every model, we used Image Augmentation with ImageDataGenerator from keras, so we can make our dataset more complete.We performed rotation, width shift, height shift, random zoom and horizontal flips to generate more images from the existing dataset.In order to use the images we already have, we employed flow\_from\_directory which allows us to read the images directly from the directory and augment them while the CNN is learning on the training data.We compiled every model with the Adam optimizer and trained them on our dataset as well with 10 epochs.After this process, some of the best models were the following:

\begin{table}[htbp]
\caption{Accuracy of each pre-trained model}
\begin{center}
\begin{tabular}{|c|c|c|}
\hline
\textbf{Pre-trained}&\multicolumn{2}{|c|}{\textbf{Accuracy of the best epoch}} \\
\cline{2-3} 
\textbf{Model} & \textbf{\textit{accuracy}}& \textbf{\textit{val\_accuracy}} \\
\hline
Xception& $\approx88\%$&$\approx94\%$ \\
\hline
ResNet50& \textbf{$\approx86\%$} & \textbf{$\approx95\%$} \\
\hline
EfficientNetV2L& $\approx87\%$& $\approx90\%$ \\
\hline
VGG16& $\approx93\%$& $\approx94\%$ \\
\hline
VGG19& \boldmath$\approx92\%$& \boldmath$\approx95\%$ \\
\hline
NasNetLarge& $\approx88\%$& $\approx84\%$ \\
\hline
\end{tabular}
\label{tab2}
\end{center}
\end{table}

The best model out of the 8 models that we trained was the VGG19 which was created by Karen Simonyan and Andrew Zisserman at the annual Imagenet Large Scale Visual Recognition Challenge competition.This model shares almost the same architecture as LeNet5 \cite{b19} which was one of the first CNN models.Now we will analyze the process behind the training and the architecture of the model.\\
As we explained in Theory, our input is pixels of an image.The images in the dataset are fixed-size of $350x350$ RGB.For each color channel, each image has $350x350$ pixel values.We can represent each tensor with dimensions $(350,350,3)$, lets assume this is $X$.Each pixel can fluctuate between values from $0$-$255$.We can represent the tensor $X$ as $X = X_{(i,j,k)}$(Fig. 8), for example, if $k = 1$ that means it exists in the Red channel.
\begin{figure}[htpb]
\centerline{\includegraphics[width=0.45\textwidth]{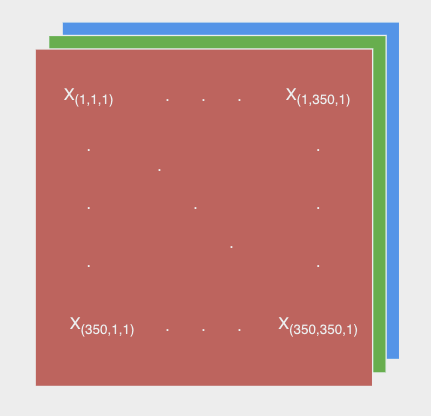}}
\caption{Representation of the input as a tensor.}
\label{fig8}
\end{figure}
The output $\hat{y}$ will be a vector of size $2x1$ as we are performing binary classification. $\hat{y} = 
\begin{bmatrix}
y_0\\
y_1
\end{bmatrix}$
with probability $p_0$ and $p_1$ respectively.Kindly note that, due to application of binary classification, the utilization of the sigmoid activation function is warranted.\\
VGG19 is a model that contains 24 layers.Out of those, only 19 of them are trainable, thus,
\begin{figure}[htpb]
\centerline{\includegraphics[width=0.45\textwidth]{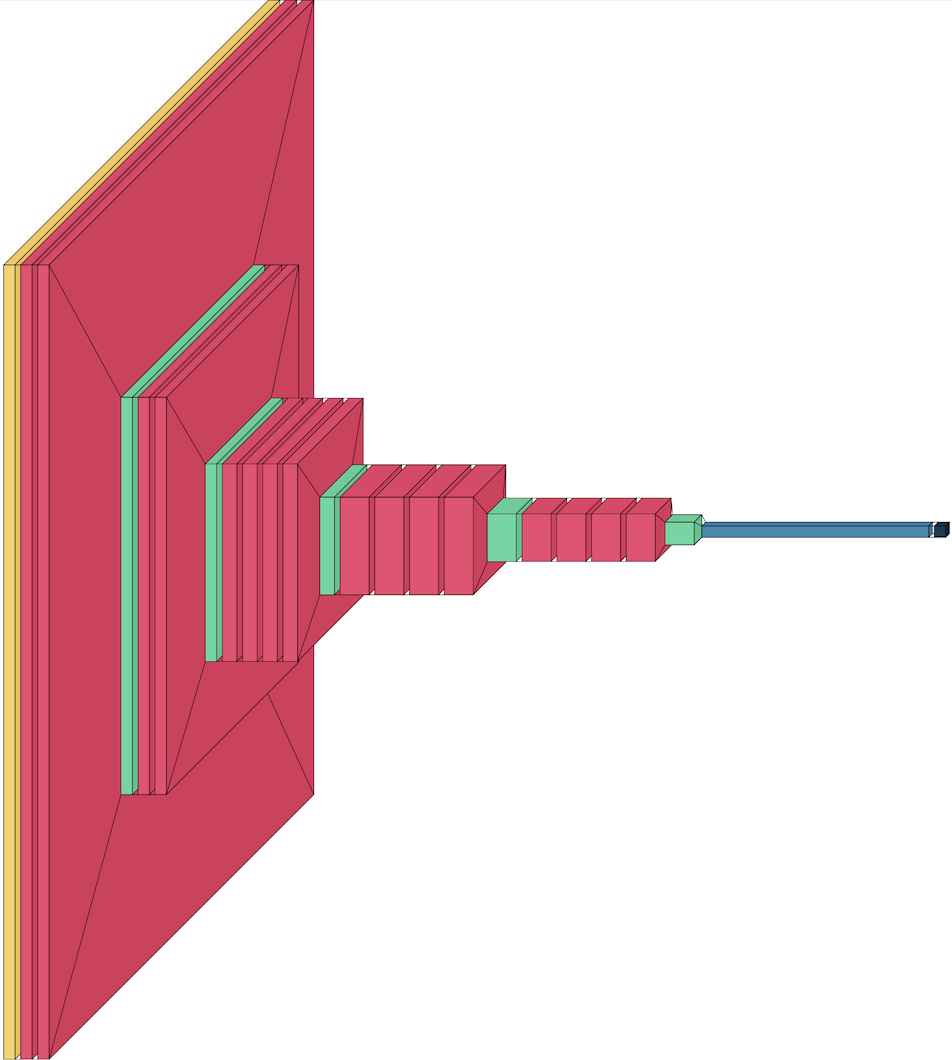}}
\caption{Architecture of VGG19.}
\label{fig9}
\end{figure}
out of 20.126.786 total parameters, only 102.402 of them are trainable, saving us a ton of computational time.In Fig. 9 we can see the architecture of the model.VGG19 contains an Input layer(with Yellow color), Dense layers with ReLU as an activation function(With Red color), Max Pooling layers(With Green color), a Flatten layer(With Blue color) and lastly, a Dense layer with the sigmoid activation function(With dark Blue color).VGG19 uses Max Pooling layers that we explained in Theory.We will deconstruct what happens in the 2D Max Pooling layer of the model.Let's take the transition of the second Dense Layer $a^{[2]}$ with the first Max-Pool layer $m^{[2]}$.The dimensions of $a^{[2]}$ are $(350,350,64)$ and the dimensions of the output $m^{[2]}$ are $(175,175,64)$.The model use $(5)$ and $(6)$ to compute the appropriate size of the filter so we can have the desired dimensions in the output.After the other blocks of Dense and Max-Pooling layers, we added a Flatten layer which takes as an input the output of the final Max-Pooling layer $m^{[16]}$ and flatten it to get as an output a dimension of $(51.200,1)$.This is because the Flatten layer takes the dimensions of the input and multiply them to get the desired result, in our case $m^{[16]}$ have dimensions $(10,10,512)$, so the output will be $10 \cdot 10 \cdot 512 = 51.200$.Finally we insert a Dense layer with a Sigmoid activation function.We compiled the model using the Adam optimizer with a learning rate of $10^{-4}$ and categorical cross-entropy as a loss function.The data was splitted into batches of 16 for both the Train and Test images using flow\_from\_directory from keras.We trained the model with 10 epochs saving the best version each time. In Fig. 10, we can see the train and the test accuracy of each epoch.
\begin{figure}[htpb]
\centerline{\includegraphics[width=0.45\textwidth]{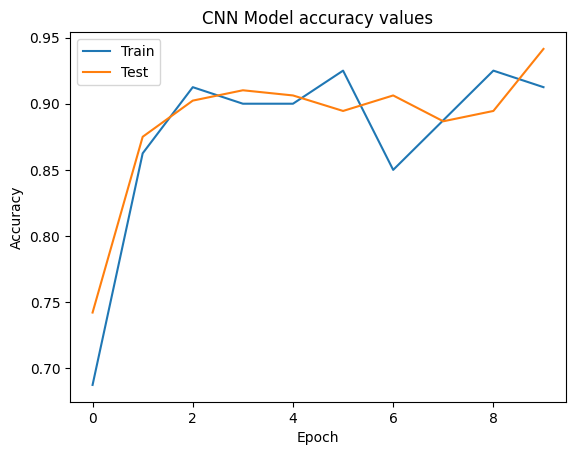}}
\caption{Accuracy values of VGG19.}
\label{fig10}
\end{figure}
Note that VGG19 is already trained on the ImageNet dataset, which contains similar images, thus, the model might get the best results in any step and the loss will not always decrease over the epochs.VGG19 managed to score a very high accuracy score with $\approx 95\%$, making it reliable enough for usage.

\section{Conclusion}
The whole point of performing this classification, as we explained in Introduction, was to optimize the FWI prediction system, taking advantage of both meteorological and geospatial data (Fig. 11)
\begin{figure}[htpb]
\centerline{\includegraphics[width=0.5\textwidth]{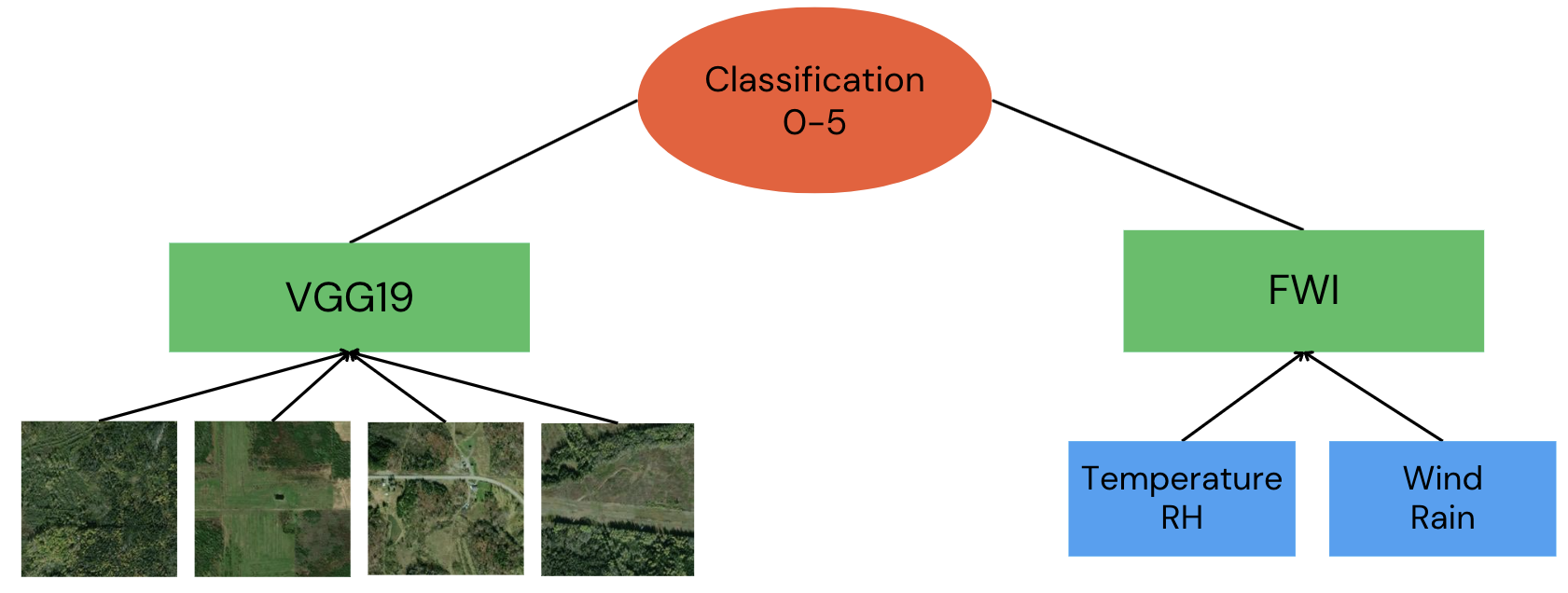}}
\caption{Final classification process.}
\label{fig11}
\end{figure}
With this technique we can compute the probability and perform a final classification of the danger level.\\
In essence, our study delves into the intricate dynamics behind wildfires, revealing the predominant role of human actions intertwined with environmental factors in their emergence. Beyond mere weather conditions, human-induced causes, such as mismanaged land practices and negligence, stand as primary contributors to these catastrophic events\cite{b20}. Employing advanced CNNs, particularly the VGG19 model, our analysis of geospatial data has emerged as a potent avenue for precise wildfire prediction and accurate identification of burnt areas within forests and rural landscapes.Central to our findings is the resounding call for collective responsibility. Each individual's proactive engagement in conserving natural habitats assumes paramount importance, given the far-reaching consequences of wildfires on both local communities and the broader environment. Safeguarding our ecosystems is no longer a choice but an imperative, safeguarding not only the present but also securing a sustainable future for generations to come.

\section{Discussion \& further work}
Transfer learning is useful for every image classification that we want to perform in datasets that include images of nature.Almost every net we implemented showed excellent results, in contrast with algorithms on meteorological data.The pre-trained models that keras offers us are generally big and can handle a huge amount of parameters, for example, VGG19 can take up to 143.6M parameters.Thus, our dataset is a bit small to be used for such a model, though transfer learning can train deep neural networks with comparatively little data.In conclusion, working with a bigger dataset that includes photographs from more biomes would've surely gave us better results.\\
One improvement we can suggest for the model is to include more classes about the damage of the forest(i.e. 0-5 with 0 having no damage from wildfires and 5 being completely destroyed).If a forest is completely destroyed(Severity = 5) then the danger level will drop dramatically, the only thing we have to take care about is regeneration concerns.One way to do that is by splitting an aerial photograph of a forest into $350x350$ images(Fig. 12).
\begin{figure}[htpb]
\centerline{\includegraphics[width=0.5\textwidth]{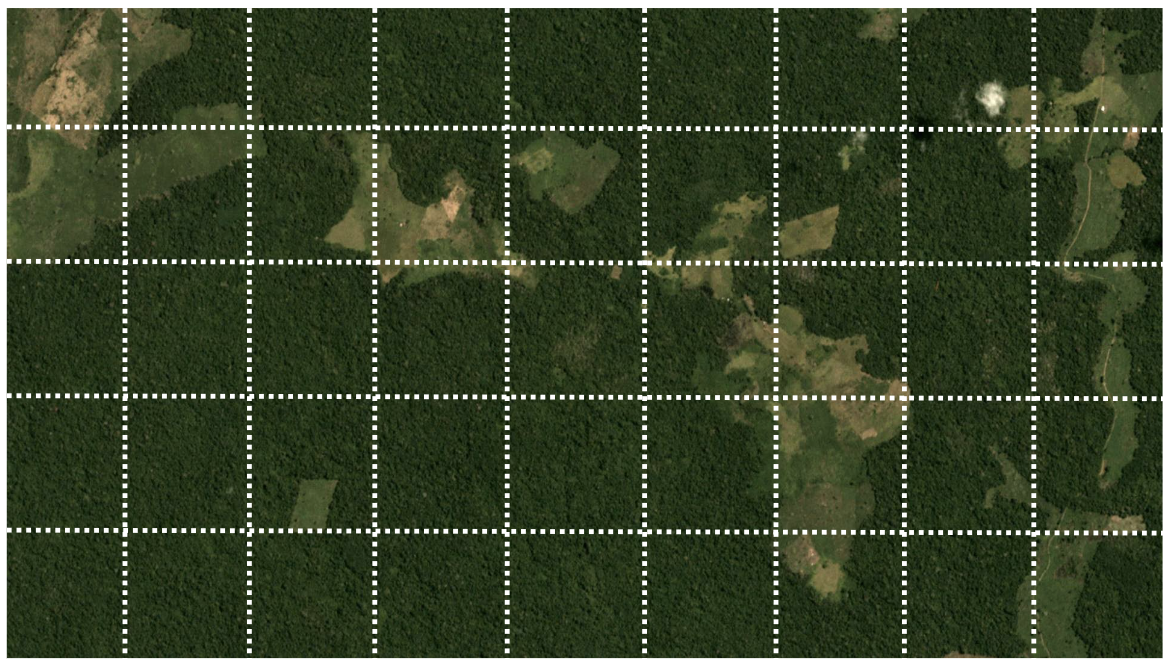}}
\caption{Satellite image of a forest splitted in $350x350$ images.}
\label{fig12}
\end{figure}
The improved model will have an output of $\hat{y} = \begin{bmatrix}
y_0\\
y_1\\
y_2\\
y_3\\
y_4\\
\end{bmatrix}$
and we have to compute the probability of a wildfire $x$ having an output $\hat{w}$ from the FWI prediction system and $\hat{y}$ from VGG19.
\[ 
p(x|\{\hat{y},\hat{w}\})
\]
Kindly note that performing multi-classification using these models is a huge advantage as pre-trained models are often used for these kind of problems($\geq1000$ classes) and they give us better results than custom nets.

\end{document}